\documentclass[10pt,twocolumn,letterpaper]{article}

\usepackage{cvpr}
\usepackage{times}
\usepackage{epsfig}
\usepackage{graphicx}
\usepackage{amsmath}
\usepackage{amssymb}
\usepackage{mathtools}
\usepackage{array}
\usepackage{multirow}
\usepackage[inline]{enumitem}
\usepackage{subfig}
\usepackage{caption}

\usepackage[pagebackref=true,breaklinks=true,letterpaper=true,colorlinks,bookmarks=false]{hyperref}

\cvprfinalcopy 

\def\cvprPaperID{1254} 

\ifcvprfinal\fi
\begin{document}

\title{AdaScan: Adaptive Scan Pooling in Deep Convolutional Neural Networks \\
for Human Action Recognition in Videos}

\author{Amlan Kar$^{1,}$\thanks{Amlan Kar and Nishant Rai contributed equally to this work.} \hspace{2em}
	Nishant Rai$^{1,}\footnotemark[1]$ \hspace{2em}
	Karan Sikka$^{2,3,}$\thanks{Part of the work was done when Karan Sikka was with UCSD.
    \texttt{karan.sikka@sri.com}} \hspace{2em}
	Gaurav Sharma$^{1}$ \vspace{0.2em}\\
	$^1$IIT Kanpur\thanks{\texttt{\{amlan, nishantr, grv\}@cse.iitk.ac.in}}\hspace{2.7em} $^2$SRI International  \hspace{2em} $^3$UCSD
}

\maketitle
\def\etal{et al\onedot}
\def\etc{etc\onedot}
\def\ie{i.e\onedot}
\def\eg{e.g\onedot}
\def\cf{cf\onedot}
\def\vs{vs\onedot}
\def\pd{\partial}
\def\grad{\nabla}
\def\Li{\mathcal{L}}
\def\O{\mathcal{O}}
\def\N{\mathbb{N}}
\def\C{\mathcal{C}}
\def\Lt{\tilde{L}}
\def\R{\mathbb{R}}
\def\X{\mathcal{X}}
\def\I{\mathcal{I}}
\def\F{\mathcal{F}}
\def\w{\textbf{w}}
\def\x{\textbf{x}}
\def\xp{\bar{\textbf{x}}}
\def\k{\textbf{k}}
\def\k{\textbf{k}}
\def\d{\boldsymbol{\delta}}
\def\y{\textbf{y}}
\def\l{\boldsymbol{\ell}}
\def\wrt{w.r.t\onedot}
\def\a{\boldsymbol{\alpha}}
\def\vertspace{0.6em}
\def\adascan{\texttt{AdaScan} }

\newcommand{\red}[1]{\textcolor{red}{#1}}

\def\algorithmautorefname{Algorithm}
\def\figureautorefname{Figure}
\def\tableautorefname{Table}
\def\equationautorefname{Eq.}
\def\sectionautorefname{Section}
\def\subsectionautorefname{Section}
\def\subsubsectionautorefname{Section}

\begin{abstract}

We propose a novel method for temporally pooling frames in a video for the task of human action
recognition. The method is motivated by the observation that there are only a small number of frames
which, together, contain sufficient information to discriminate an action class present in a video,
from the rest. The proposed method learns to pool such discriminative and informative frames, while
discarding a majority of the non-informative frames in a single temporal scan of the video. Our
algorithm does so by continuously predicting the discriminative importance of each video frame and
subsequently pooling them in a deep learning framework. We show the effectiveness of our proposed
pooling method on standard benchmarks where it consistently improves on baseline pooling methods,
with both RGB and optical flow based Convolutional networks. Further, in combination with
complementary video representations, we show results that are competitive with respect to the
state-of-the-art results on two challenging and publicly available benchmark datasets. 

\end{abstract}

\section{Introduction}

Rapid increase in the number of digital cameras, notably in cellphones, and cheap internet with high
data speeds, has resulted in a massive increase in the number of videos uploaded onto the internet
\cite{cisco}.
Most of such videos, \eg on social networking websites, have humans as their central subjects.
Automatically predicting the semantic content of videos, \eg the action the human is performing,
thus, becomes highly relevant for searching and indexing in this fast growing database.  In order to
perform action recognition in such videos, algorithms are required that are both easy and fast to
train and, at the same time, are robust to noise, given the real world nature of such videos.

\begin{figure}
\centering
\includegraphics[width=\columnwidth]{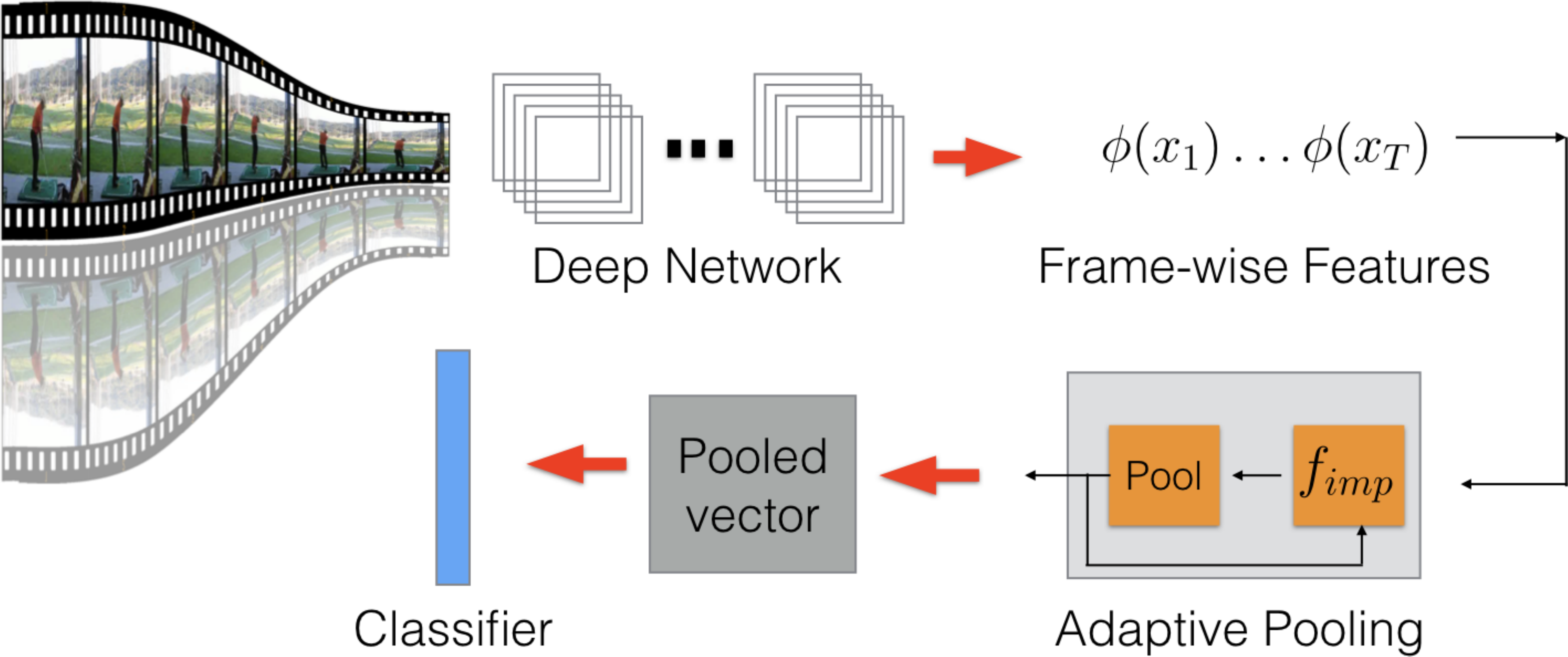} 
\includegraphics[width=0.9\columnwidth, trim= 585 200 940 200, clip]{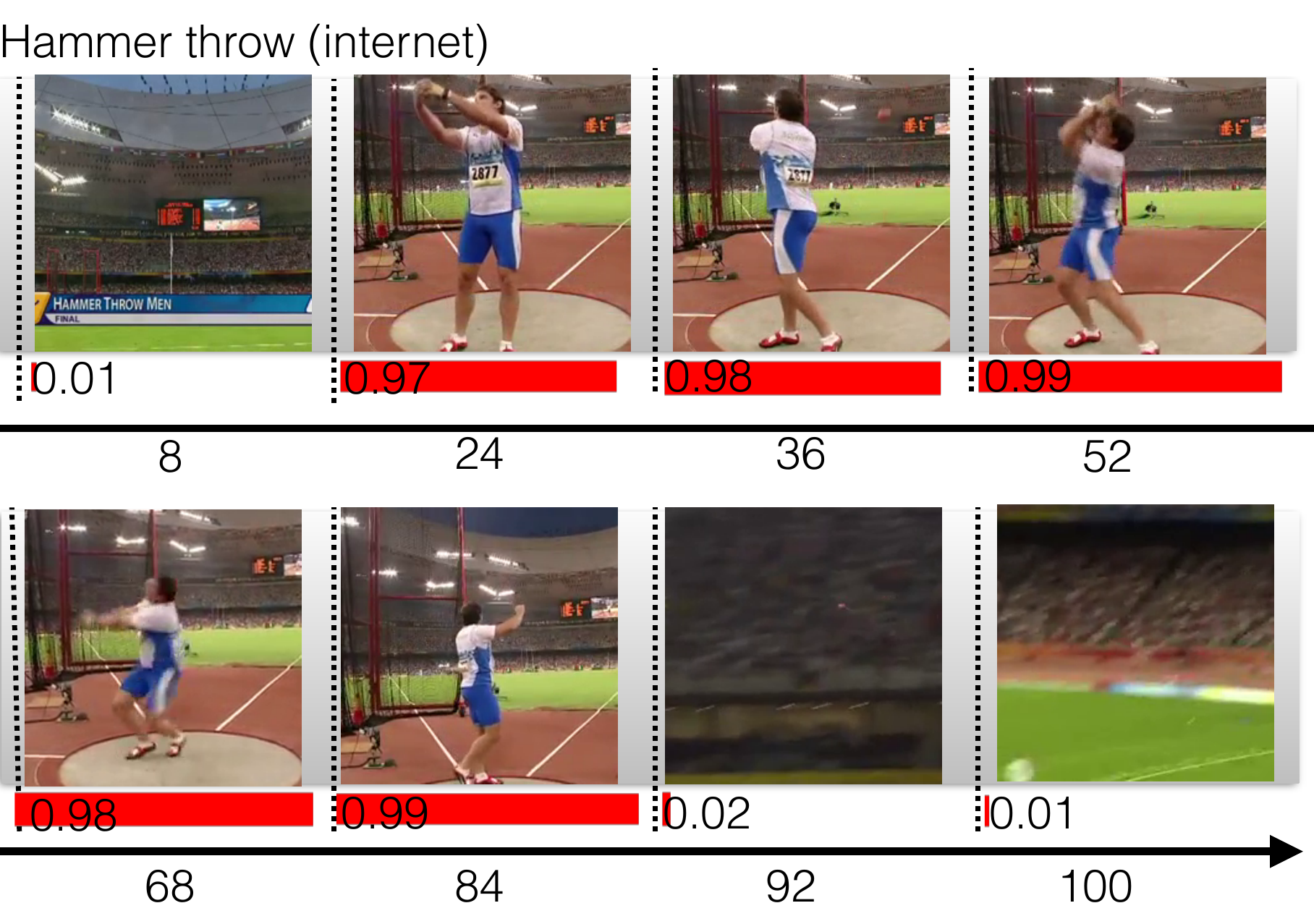}
\vspace{-0.8em}
\caption[Illus ]{
        (Top) Illustration of proposed \adascan. It first extracts deep features for each frame in a
        video and then passes them to the proposed Adaptive Pooling module, which recursively pools
        them while taking into account their discriminative importances---which are predicted inside
        the network. The final pooled vector is then used for classification. (Bottom) Predicted
        discriminative importance for a video that was downloaded from the internet\footnotemark and
        ran through \adascan trained on UCF101. The numbers and bars on the bottom indicate the
        predicted importance $\in [0, 1]$ and the timeline gives the relative frame position in
        percentile (see \autoref{Qualsec}).
}
\label{figIllus}
\end{figure}

A popular framework for performing human action recognition in videos is using a temporal pooling
operation to `squash' the information from different frames in a video into a summary vector.
\footnotetext{Video downloaded from \url{https://www.youtube.com/watch?v=KnHUAc20WEU} and cropped
        from $3$--$18$ seconds} Mean and max pooling, \ie taking the average or the coordinatewise
max of the (features of the) frames, are popular choices, both with classic `shallow' as well as
recent `deep' methods \cite{SimonyanNIPS2014, wangIJCV2015, laptevCVPR2008, wang2015action}.
However, these pooling methods consider all frames equally and are not robust to noise, \ie to the
presence of  video frames that do not correspond to the target action \cite{niebles2010modeling,
        Fernando2016,bilen2016dynamic, li2013dynamic, Wang-Hoai-CVPR16, zhu2016key}. This
results in a loss in performance as noted by many host algorithms, with both shallow and deep
pipelines \eg \cite{caba2015activitynet, Fernando2016, sikka2016discriminatively, li2013dynamic}.
Several methods have proposed solutions to circumvent the limitations of these pooling methods. Such
solutions either use Latent Variable Models \cite{niebles2010modeling, tang2012learning,
        gaidonIJCV2014, sikka2016discriminatively, liCVPR2015}, which require an additional
inference step during learning, or employ a variant of Recurrent Neural Networks (RNN)
\cite{sharma2015action, YueCVPR2015} which have intermediate hidden states that are not immediately
interpretable. In this work we propose a novel video pooling algorithm that learns to  dynamically
pool video frames for action classification, in an end-to-end learnable manner, while producing
interpretable intermediate `states'. We name our algorithm \adascan since it is able to both
adaptively pool video frames, and make class predictions in a single temporal scan of the video.  As
shown in \autoref{figIllus}, our algorithm internally predicts the discriminative importance of each frame in a
video and uses these states for pooling. The proposed algorithm is set in a weakly supervised setting for action classification
in videos, where labels are provided only at video-level and not at frame-level
\cite{niebles2010modeling, zhu2016key, sikka2016discriminatively, li2013dynamic,
        caba2015activitynet}. This problem is extremely relevant due to the difficulty and
non-scalability of obtaining frame-level labels. The problem is also very challenging as potentially
noisy and untrimmed videos may contain distractive frames that do not belong to the same action
class as the overall video.

Algorithms based on the Multiple Instance Learning (MIL) framework try to solve this problem by
alternating between spotting relevant frames in videos and (re-)learning the model. Despite
obtaining promising results, MIL is (i) prone to overfitting, and (ii) by design, fails to take into
account the contributions of multiple frames together, as noted recently
\cite{sikka2016discriminatively, liCVPR2015}. More recently, Long Short Term Memory (LSTM) networks
have also been used for video classification. They encode videos using a recurrent operation and
produce hidden vectors as the final representations of the videos \cite{sharma2015action,
YueCVPR2015, donahueCVPR2015}. Despite being able to model reasonably long-term temporal
dependencies, LSTMs are not very robust to noise and have been shown to benefit from explicit,
albeit automatic, removal of noisy frames \cite{gan2016you, zhu2016key}. The proposed algorithm does
not require such external noisy frame pruning as it does so by itself while optimizing the
classification performance in a holistic fashion.

In summary we make the following contributions. (1) We propose a novel approach for human action classification in videos
that (i) is able to identify informative frames in the video and only pool those, while discarding
others, (ii) is end-to-end trainable along with the representation of the
images, with the final objective of discriminative classification, and (iii) works in an inductive
setting, \ie given the training set it learns a parametrized function to pool novel videos
independently, without requiring the whole training set, or any subset thereof, at test time. (2) We
validate the proposed method on two challenging publicly available video benchmark datasets and show
that (i) it consistently outperforms relevant pooling baselines and (ii) obtains
state-of-the-art performance when combined with complimentary representations of videos. (3) We
also analyze qualitative results to gain insights to the proposed algorithm and show that our
algorithm achieves high performance while only pooling from a subset of the frames. 

\ \vspace{0.2em}

\section{Related Work}

Many earlier approaches relied on using a Bag of Words (BoW) based pipeline. Such methods typically
extracted local spatio-temporal features and encoded them using a dictionary \cite{laptevCVPR2008,
wang2009evaluation, dollar2005behavior, scovanner2007, perronninECCV2010, wangIJCV2015}. One of the
first works \cite{laptevCVPR2008} described a video with BoW histograms that encoded Histograms of
Gradients (HoG) and Histograms of Flow (HoF) features over 3D interest points. Later works improved
upon this pipeline in several ways \cite{peng2016bag, wang2012comparative} by using dense
sampling for feature extraction \cite{wang2009evaluation}, describing trajectories instead of 3D
points \cite{jiang2012trajectory, wang2013dense}, and using better pooling and encoding methods
\cite{wang2012comparative, perronninECCV2010, peng2016bag}. Improving upon these methods Wang \etal
\cite{wangIJCV2015} proposed the Improved Dense Trajectories (iDT) approach that showed significant
improvement over previous methods by using a combination of motion stabilized dense trajectories,
histogram based features and Fisher Vector (FV) encodings with spatio-temporal pyramids. Some recent
methods have improved upon this pipeline by either using multi-layer fisher vectors
\cite{peng2014action} or stacking them at multiple temporal scales \cite{lan2015beyond}. All of
these approaches rely on the usage of various local features combined with standard pooling operators. 

While the above methods worked with an orderless representation, another class of methods worked on explicitly
exploiting the spatial and temporal structure of human activities. Out of these, a set of methods have used latent structured SVMs for modeling the temporal structure in human
activities. These methods typically alternate between identifying discriminative frames (or
segments) in a video (inference step) and learning their model parameters.  Niebles \etal
\cite{niebles2010modeling} modeled an activity as a composition of latent temporal segments with
anchor positions that were inferred during the inference step. Tang \etal \cite{tang2012learning}
improved upon Niebles \etal by proposing a more flexible approach using a variable duration HMM
that factored each video into latent states with variable durations. Other approaches have also used
MIL and its variants to model discriminative frames in a video, with or without a temporal structure
\cite{raptis2013poselet, sikka2016discriminatively, gaidonPAMI2013, zhuICCV2013, WangICCV2013,
li2013dynamic, satkin2010modeling}. Most related to our work is the dynamic pooling appoach used by Li \etal
\cite{li2013dynamic} who used a scoring function to identify discriminative frames in a video and
then pooled over only these frames. In contrast, our method does not solve an inference
problem, and instead explicitly predicts the discriminative importance of each frame and pools
them in a single scan. Our work is also inspired by an early work by Satkin \etal
\cite{satkin2010modeling} who identified the best temporal boundary of an action, defined as the
minimum number of frames required to classify this action, and obtained a final representation by
pooling over these frames.

Despite the popularity of deep Convolutional Neural Networks (CNN) in image classification, it is
only recently that deep methods have achieved performance comparable to shallow methods for video
action classification. Early approaches used 3D convolutions for action recognition 
\cite{ji20133d, KarpathyCVPR2014}; while these showed decent results on the task, the top
performances were still obtained by the traditional non-deep methods.  Simonyan \etal \cite{SimonyanNIPS2014} proposed the two-stream
deep network that combined a spatial network (trained on RGB frames) and a temporal network (trained
on stacked flow frames) for action recognition. Ng \etal \cite{YueCVPR2015} highlighted a drawback
in the two-stream network that uses a standard image CNN instead of a
specialized network for training videos. This results in the two-stream network not being able to
capture long-term temporal information. They proposed two deep networks for action classification by
(i) adding standard temporal pooling operations in the network, and (ii) using LSTMs for feature
pooling. Recent methods have also explored the use of LSTMs for both predicting action classes
\cite{li2016videolstm, sharma2015action, srivastava2015unsupervised, li2016videolstm} and video
caption generation \cite{donahueCVPR2015, yao2015describing}. Some of these techniques have also
combined attention with LSTM to focus on specific parts of a video (generally spatially) during
state transitions \cite{li2016videolstm, sharma2015action, yao2015describing}. Our work bears
similarity to these attention based frameworks in predicting the relevance of different parts of the
data. However it differs in several aspects: (i) The attention or disriminative importance utilized
in our work is defined over temporal dimension \vs the usual spatial dimension, (ii) we predict
this importance score in an online fashion, for each frame, based on the current frame and already
pooled features, instead of predicting them together for all the frames \cite{yao2015describing},
and (iii) ours is a simple formulation that combines the prediction with standard mean pooling
operation to dynamically pool frame-wise video features. Our work is also related to LSTMs through its recursive formulation but differs in producing a clearly interpretable intermediate state along with the
importance of each frame \vs LSTM's generally non-interpretable hidden states. It is also worth mentioning the
work on Rank Pooling and Dynamic Image Networks that use a ranking function to pool a video
\cite{bilen2016dynamic, Fernando2016}. However, compared to current methods their approach
entails a non-trivial intermediate step that requires solving a ranking formulation for pooling each
vector.  

\section{Proposed Approach}
\label{approach}

We now describe the proposed approach, that we call \adascan(Adaptive Scan Pooling Network), in
detail. We denote a video as 
\begin{equation}
X = [\x_1,\ldots,\x_T], \x_t \in \R^{224 \times 224 \times K},
\end{equation} 
with each frame $\x_t$ either represented as RGB images ($K=3$), or as a stack of optical flow
images of neighbouring frames \cite{SimonyanNIPS2014} ($K=20$ in our experiments). We work in a
supervised classification setting with a training set 
\begin{equation}
\X = \{ (X_i, y_i) \}_{i=1}^N \subset \R^{224 \times 224 \times K \times T} \times \{1,\ldots,C\}, 
\end{equation} 
where $X_i$ is a training video and $y_i$ is its class label (from one of the $C$ possible classes). 
In the following, we drop the subscript $i$, wherever it is not required, for brevity.

\adascan is a deep CNN augmented with an specialized pooling module (referred to as `Adaptive
Pooling') that scans a video and dynamically pools the features of select frames to generate a final
pooled vector for the video, adapted to the given task of action classification. As shown in
\autoref{figIllus}, our model consists of three modules that are connected to each other
sequentially. These three modules serve the following purposes, respectively: (i) feature
extraction, (ii) adaptive pooling, and (iii) label prediction. The feature extractor module
comprises of all the convolutional layers along with the first fully connected (FC-6) layer of the
VGG-16 network of Simonyan \etal \cite{SimonyanICLR2015}. This module is responsible for extracting
deep features from each frame $\x_t$ of a video, resulting in a fixed dimensional vector, denoted as
$\phi(\x_t) \in \R^{4096}$. The purpose of the Adaptive Pooling module is to selectively pool the
frame features by aggregating information from only those frames that are discriminative for the
final task, while ignoring the rest. It does so by recursively predicting a score that quantifies
the discriminative importance of the current frame, based on (i) the features of the current frame,
and (ii) the pooled vector so far. It then uses this score to update the pooled vector (described
formally in the next section). This way it aggregates discriminative information only by pooling
select frames, whose indices might differ for different videos, to generate the final dynamically
pooled vector for the video. This final vector is then normalized using an $\ell_2$ normalization
layer and the class labels are (predicted) using a FC layer with softmax function. We now describe
the adaptive pooling module of \adascan in more detail and thereafter provide details regarding the
loss function and learning procedure.

\subsection{Adaptive Pooling} 
\label{adaptive_pooling}
This is the key module of the approach which dynamically pools the features of the frames of a
video. It does a temporal scan over the video and pools the frames by inferring the discriminative
importance of the current frame feature given the feature vector and the pooled video vector so
far. In the context of video classification, we want the predicted discriminative importance of a
frame to be high if the frame contains information positively correlated to the class of the video,
and possibly negatively correlated to the rest of the classes,  and low if the frame is either
redundant, \wrt already pooled frames, or does not contain any useful information for the
classification task. We note that this definition of importance is similar to the notion of
discriminativeness of a particular part of the data as used in prior MIL based methods.  However,
contrary to MIL based methods, which effectively weight the frames with a one-hot vector, our
algorithm is naturally able to focus on more than one frame in a video, if required, while
explicitly outputting the importances of all the frames in an online fashion. 

Let us denote the adaptive pooled vector till the initial $t$ frames for a video $X$ as $\psi(X, t)$.
The aim is now to compute the vector after pooling all the $T$ frames in a video \ie
$\psi(X, T)$.  
The Adaptive Pooling module implements the pooling by recursively computing two operations. The
first operation, denoted as $f_{imp}$, predicts the discriminative importance, $\gamma_{t+1} \in [0,
1]$, for the next \ie $(t+1)^{th}$ frame given its CNN feature, $\phi(\x_{t+1})$, and the pooled features till time $t$, $\psi(X, t)$. We denote the importance scores of the frames of a video as a sequence of
reals $\Gamma = \{\gamma_1, \ldots, \gamma_T\} \in [0,1]$.  The second operation is a weighted
mean pooling operation that calculates the new pooled features $\psi(X, t+1)$ by
aggregating the previously pooled features with the features from current frame and its predicted
importance. The operations are formulated as:
\begin{align}
        \gamma_{t+1} &= f_{imp}(\psi(X, t), \phi(\x_{t+1}))\\
        \psi(X, t+1) &= \frac{1} {\hat{\gamma}_{t+1}} \left( \hat{\gamma}_t 
                            \psi(X, t) + \gamma_{t+1} \phi(\x_{t+1})\right) \\
  \textrm{where, \ \ } \hat{\gamma}_p & = \sum_{k=1}^p \gamma_k
\end{align}

Effectively, at $t^{th}$ step the above operation does a \emph{weighted} mean pooling of all the
frames of a video, with the weights of the frame features being the predicted discriminative
importance scores $\gamma_1, \ldots, \gamma_t$. 

We implement the attention prediction function $f_{imp}(\cdot)$ as a Multilayer Perceptron (MLP)
with three layers. As the underlying operations for $f_{imp}(\cdot)$ rely only on standard linear
and non-linear operations, they are both fast to compute and can be incorporated easily inside a CNN
network for end-to-end learning. In order for $f_{imp}(\cdot)$ to consider both the importance and
non-redundancy of a frame we feed the difference between the current pooled features and features
from the next frame to the Adaptive Pooling module.
We found this simple modification, of feeding the difference, to not only help reject redundant
frames but also improve generalization. We believe this is due to the fact that the residual might
be allowing the Adaptive Pooling module to explicitly focus on unseen features while making a
decision on whether to pool them (additively) or not. 


Owing to its design, our algorithm is able to maintain the simplicity of a mean pooling operation
while predicting and adapting to the content of each incoming frame. Moreover at every timestep we
can easily interpret both the discriminative importance and the pooled vector for a video, leading
to an immediate extension to an online/streaming setting, which is not the case for most recent
methods. 

\subsection{Loss Function and Learning}
\label{secLoss}

We formulate the loss function using a standard cross entropy loss $\mathcal{L}_{CE}$ between the
predicted and true labels. In order to direct the model towards selecting few frames from a video, we
add an entropy based regularizer $\mathcal{L}_E$ over the predicted scores, making the full objective as
\begin{align}
        \mathcal{L}(X, y) &= \mathcal{L}_{CE}(X, y) + \lambda \mathcal{L}_{E}(\Gamma)\\
        \mathcal{L}_{E}(\Gamma) &= -\sum_k \frac{e^{\gamma_k}}{N} \log
        \left(\frac{e^{\gamma_k}}{N}\right) \\
       \gamma_k, \lambda &\geq 0, N = \sum_t e^{\gamma_t}
\end{align}
The regularizer minimizes the entropy over the normalized(using softmax) discriminative scores. Such a regularizer
encourages a peaky distribution of the importances, \ie it helps select only the discriminative
frames and discard the non discriminative ones when used with a discriminative loss. We also
experimented with the popular sparsity promoting $\ell_1$ regularizer, but found it to be too
aggressive as it led to selection of very few frames, which adversely affected the performance.
The parameter $\lambda$ is a trade-off parameter which balances between a sparse selection of frames
and better minimization of the cross entropy classification loss term. If we set $\lambda$ to
relatively high values we expect fewer number of frames being selected, which would make the
classification task harder \eg single frame per video would make it same as image classification.
While, if the value of $\lambda$ is relatively low, the model is expected to select larger number of
frames and also possibly overfit.
We show empirical results with varying $\lambda$ in the experimental \autoref{secExpLambda}.

\section{Experimental Results} 

We empirically evaluate our approach on two challenging publicly available human action
classification datasets. We first briefly describe these datasets, along with their experimental
protocol and evaluation metrics. We then provide information regarding implementation of our work.
Thereafter we compare our algorithm with popular competetive baseline methods. We also study the
effect of the regularization used in \adascan and compare our approach with previous
state-of-the-art methods on the two datasets. We finally discuss qualitative results to provide
important insights to the proposed method. 
\vspace{0.8em} \\
\textbf{HMDB51\footnote{\url{http://serre-lab.clps.brown.edu/resource/hmdb-a-large-human-motion-database/}}}
\cite{kuehneICCV11} dataset contains around $6800$ video clips from $51$ action classes. These
action classes cover wide range of actions -- facial actions, facial action with object
manipulations, general body movement, and general body movements with human interactions. This
dataset is challenging as it contains many poor quality video with significant camera motions and
also the number of samples are not enough to effectively train a deep network
\cite{SimonyanNIPS2014, wang2015towards}.  We report classification accuracy for $51$ classes across
$3$ splits provided by the authors \cite{kuehneICCV11}. 
\vspace{0.8em} \\ \textbf{UCF101\footnote{\url{http://crcv.ucf.edu/data/UCF101.php}}}
\cite{soomro2012ucf101} dataset contains $13320$ videos from $101$ action classes that are divided
into $5$ categories- human-object interaction, body-movement only, human-human interaction, playing
musical instruments and sports.  Action classification in this datasets is challenging owing to
variations in pose, camera motion, viewpoint and spatio-temporal extent of an action. Owing to these
challenges and higher number of samples this dataset is often used for evaluation in previous works.
We report classification accuracy for $101$ classes across the $3$ train/test splits provided
by the authors \cite{soomro2012ucf101}.   

\subsection{Implementation Details}
To implement \adascan, we follow Simonyan \etal \cite{SimonyanNIPS2014}, and use a two-stream
network that consists of a spatial and a temporal $16$ layer VGG network \cite{SimonyanICLR2015}. We
generate a $20$ channel optical flow input, for the temporal network, by stacking both X and Y
direction optical flows from $5$ neighbouring frames in both directions \cite{SimonyanNIPS2014,
wang2015towards}. We extract the optical flow using the
tool\footnote{\url{https://github.com/wanglimin/dense_flow}} provided by Wang \etal
\cite{wang2015towards}, that uses TV-L1 algorithm and discretizes the optical flow fields in the
range of $[0, 255]$ by a linear transformation. As described in \autoref{approach} our network trains
on a input video containing multiple frames instead of a single frame as was done in the two-stream
network \cite{SimonyanNIPS2014}.  Since videos vary in the number of frames and fitting an entire
video on a standard GPU is not possible in all cases, we prepare our input by uniformly sampling
$25$ frames from each video. We augment our training data by following the multiscale cropping
technique suggested by \cite{wang2015towards}. For testing, we use $5$ random samples of $25$ frames
extracted from the video, and use $5$ crops of $224\times 224$ along with their flipped versions. We
take the mean of these predictions for the final prediction for a sample.

We implement the Adaptive Pooling layer's $f_{imp}(\cdot)$ function, as described in
\autoref{approach}, using a three layer MLP with $\tanh$ non linearities and sigmoid activation at
the final layer. We set the initial state of
the pooled vector to be same as the features of the first frame. We found this initialization to be
stable as compared to initialization with a random vector. We initialize the components of the
Adaptive Pooling module using initialization proposed by Glorot \etal
\cite{glorot2010understanding}. We also found using the residual of the pooled and current frame
vector as input to the Adaptive Pooling module to work better than their concatenation. 

We initialize the spatial network for training UCF101 from VGG-16 model \cite{SimonyanICLR2015}
trained on ImageNet \cite{DengCVPR2009}. For training the temporal network on UCF101, we initialize
its convolutional layers with the $16000$ iteration snapshot provided by Wang \etal
\cite{wang2015towards}. For training HMDB51 we initialize both the spatial and temporal network by
borrowing the convolutional layer weights from the corresponding network trained on UCF101. During
experiments we observed that reinitializing the Adaptive Pooling module randomly performed better
than initializing with the weights from the network trained on UCF101. We also tried initializing
the network trained on HMDB51 with the snapshot provided by \cite{wang2015towards} and with an
ImageNet pre-trained model but found their performance to be worse. Interestingly, from the two
other trials, the model initialized with ImageNet performed better, showing that training on
individual frames for video classification might lead to less generic features due to the noise
injected by the irrelevant frames for an action class. We found it extremely important to use
separate learning rates for training the Adaptive Pooling module and fine-tuning the Convolutional
layers. We use the Adam solver\cite{kingma2014adam} with learning rates set to $1e-3$ for the
Adaptive Pooling module and $1e-6$ for the Convolutional layers. We use dropout with
high (drop) probabilities ($=0.8$) both after the FC-6 layer and the Adaptive Pooling module and found
it essential for training. We run the training for $6$ epochs for the spatial network on both
datasets. We train the temporal network, for $2$ epochs on UCF101 and $6$ epochs on HMDB51. We
implement our network using the tensorflow toolkit\footnote{https://www.tensorflow.org}. 
\vspace{0.8em} \\ \textbf{Baselines and complementary features.} For a fair comparison with standard
pooling approaches, we implement three baselines methods using the same deep network as \adascan
with end-to-end learning. We implement mean and max pooling by replacing the Adaptive Pooling module
with mean and max operations. For implementing MIL, we first compute classwise scores for each frame
in a video and then take a max over the classwise scores across all the frames prior to the softmax
layer. For complimentary features we compute results with improved dense trajectories (iDT)
\cite{wangIJCV2015} and 3D convolutional (C3D) features \cite{tran2015learning} and report
performance using weighted late fusion.  We extract the iDT features using the executables provided
by the authors \cite{wangIJCV2015} and use human bounding boxes for HMDB51 but not for UCF101. We
extract FV for both datasets using the implementation provided by Chen \etal \cite{sun2013large}.
For each low-level feature\footnote{Trajectory, HOG, HOF, Motion Boundary Histograms (X and Y)},
their implementation first uses Principal Component Analysis (PCA) to reduce the dimensionality to
half and then trains a Gaussian Mixture Models (GMM).  The GMM dictionaries, of size $512$, are used
to extract FV by using the \texttt{vlfeat} library \cite{vedaldi08vlfeat}. The final FV is formed by
applying both power normalization and $\ell_2$ normalization to per features FV and concatenating
them.  Although Chen \etal have only provided the GMMs and PCA matrices for UCF101, we also use them
for extracting FVs for HMDB51. For computing C3D features we use the Caffe implementation provided
by Tran \etal \cite{tran2015learning} and extract features from the FC-6 layer over a $16$ frame
window. We compute final feature for each video by max pooling all the features followed by $\ell_2$
normalization.

\begin{table}[t]
    \centering
    \begin{tabular}{ccccc}
	    \hline 
        Network &	Max Pool& MIL & Mean Pool& \adascan\\ 
	    \hline \hline
        Spatial & $77.2$ & $76.7$ & $78.0$ & $\mathbf{79.1}$\\ 
        Temporal & $80.3$ & $79.1$ & $80.8$ &$\mathbf{81.7}$ \\ \hline
	\end{tabular}
    \vspace{-1em}
    \caption{Comparison with baselines on UCF101 - Split 1 in terms of multiclass classification
            accuracies.}
    \label{tabSplit1}
    \vspace{-1em}
\end{table}

\subsection{Quantitative Results}

\subsubsection{Comparison with Pooling Methods} \autoref{tabSplit1} gives the performances of
\adascan along with three other commonly used pooling methods as baselines \ie max pooling
(coordinate-wise $\max$), MIL (multiple instance learning) and mean pooling, on the Split 1 of the
UCF101 dataset. MIL is the weakest, followed by max pooling and then mean pooling ($76.7, 77.2, 78.0$
resp.\ for spatial network and $79.1, 80.3, 80.8$ for the temporal one), while the proposed \adascan
does the best ($79.1$ and $81.7$ for spatial and temporal networks resp.). The trends observed here
were typical --- we observed that, with our implementations, among the three baselines, mean pooling
was consistently performing better on different settings. This could be the case since MIL is known
to overfit as a result on focussing only on a single frame in a video
\cite{sikka2016discriminatively, liCVPR2015}, while max pooling seems to fail to summarize
relevant parts of an actions (and thus overfit) \cite{Fernando2016}. 
Hence, in the following experiments we mainly compare with mean pooling.

\begin{table}[t]
        \centering
        {
                \begin{tabular}{r|cc|cc}
                       \hline 
                        \multicolumn{1}{c|}{}& \multicolumn{2}{c|}{Spatial network} &
                                              \multicolumn{2}{c}{Temporal network}\\ 
                        \small{Split} & \small{Mean Pool} & \small{\adascan} & \small{Mean Pool} & \small{\adascan} \\ \hline \hline
                        1   & $78.0$ & $\mathbf{79.1}$ & $80.8$ & $\mathbf{82.3}$ \\ 
                        2   & $77.2$ & $\mathbf{78.2}$ & $82.7$ & $\mathbf{84.1}$ \\ 
                        3   & $77.4$ & $\mathbf{78.4}$ & $83.7$ & $        83.7 $ \\ \hline
                        \small{Avg} & $77.6$ & $\mathbf{78.6}$ & $82.4$ & $\mathbf{83.4}$ \\ \hline 
					   \multicolumn{5}{c}{UCF101 \cite{soomro2012ucf101} \vspace{0.2em}} \\                 
                \end{tabular}        
        }
        {
                \begin{tabular}{r|cc|cc}
                        \multicolumn{5}{c}{ } \\ 
                      \hline 
                        \multicolumn{1}{c|}{}& \multicolumn{2}{c|}{Spatial network} &
                                              \multicolumn{2}{c}{Temporal network}\\ 
                        \small{Split} & \small{Mean Pool} & \small{\adascan} & \small{Mean Pool} & \small{\adascan} \\ \hline \hline
                        1   & $41.3$ & $\mathbf{41.8}$ & $48.8$ & $\mathbf{49.3}$ \\ 
                        2   & $40.3$ & $\mathbf{41.0}$ & $48.8$ & $\mathbf{49.8}$ \\ 
                        3   & $41.3$ & $\mathbf{41.4}$ & $48.3$ & $\mathbf{48.5}$ \\ \hline 
                        \small{Avg} & $40.9$ & $\mathbf{41.4}$ & $48.6$ & $\mathbf{49.2}$ \\ \hline 
                          \multicolumn{5}{c}{HMDB51 \cite{kuehneICCV11} \vspace{0.2em}} \\
                \end{tabular}
                \vspace{-1em}
        }
        \caption{Comparison of \adascan with mean pooling. We report multiclass classification
                accuracies.}
        \label{tabComp}
    \vspace{-1em}
\end{table}
\vspace{-1em}

\subsubsection{Detailed Comparison with Mean Pooling}

\begin{figure*}

\begin{minipage}{0.65\textwidth}
    \resizebox{\linewidth}{!} 
    {
    \begin{tabular}{l|ccccc|cc}
        \hline	
        & Two- & Very && & Add. & & \\ 
        Method & \small{stream} & deep & \small{LSTM}  & Attn  & Opti.& UCF101 & HMDB51 \\ \hline \hline
        Simonyan \etal\cite{SimonyanNIPS2014} & \checkmark &  &  &  &  & $88.0$ & $59.4$  \\
        Wang \etal\cite{wang2015towards} & \checkmark &   &  &  &  & $88.0$ & $59.4$  \\ 
        Yue \etal \cite{YueCVPR2015} & \checkmark & \checkmark  &  &  &  & $88.2$ & --  \\ 
        Yue \etal \cite{YueCVPR2015} & \checkmark & \checkmark  & \checkmark &  &  & $88.6$ & --  \\ 
        Wang \etal \cite{wang2015action} & \checkmark & \checkmark  &  &  &  & $90.3$ & $63.2$  \\ 
        Sharma \etal \cite{sharma2015action} & &\checkmark & \checkmark & \checkmark & & $77.0^*$ & $41.3$ \\
        Li \etal \cite{li2016videolstm} & \checkmark & \checkmark  & \checkmark & \checkmark & & $89.2$ & $56.4$  \\ 
        Bilen \etal \cite{bilen2016dynamic} &  &  &  &  & \checkmark & $89.1$ & $65.2$  \\ 
        Wang \etal \cite{Wang_Transformation} & \checkmark & \checkmark  &  &  & \checkmark & $92.4$ & $62.0$ \\ 
        Zhu  \etal \cite{zhu2016key} & \checkmark & \checkmark  &  &  & \checkmark & $93.1$ & $63.3$ \\
		Wang \etal \cite{wang2016temporal} & \checkmark & \checkmark  &  &  &  & $94.2$ & $69.4$ \\ \hline
        Tran \etal \cite{tran2015learning} & \multicolumn{5}{c|}{3D convolutional filters} & $83.4$ & $53.9$ \\ 
        iDT \cite{wangIJCV2015} & \multicolumn{5}{c|}{shallow} & $84.3$ & $58.4$ \\ 
        MIFS \cite{lan2015beyond} & \multicolumn{5}{c|}{shallow} & $88.5$ & $63.8$ \\ \hline 
        \adascan & \checkmark & \checkmark &  &  &  & $89.4$ & $54.9$ \\ 
        \ \ + iDT & \multicolumn{5}{c|}{late fusion} & $91.3$ & $61.0$\\ 
        \ \ + iDT + C3D& \multicolumn{5}{c|}{late fusion} & $93.2$ & $66.9$\\ \hline
    \end{tabular}
    }
    \captionof{table}{Comparison with existing methods (Attn. -- Spatial Attention, Add. Opti. --
             Additional Optimization). ($^*$ Results are as reported by \cite{li2016videolstm})
             }
    \label{soa1}
\end{minipage}
\hfill 
\begin{minipage}{0.33\textwidth}
    \resizebox{\linewidth}{!} 
    {
	\includegraphics[trim=30 10 140 10,clip]{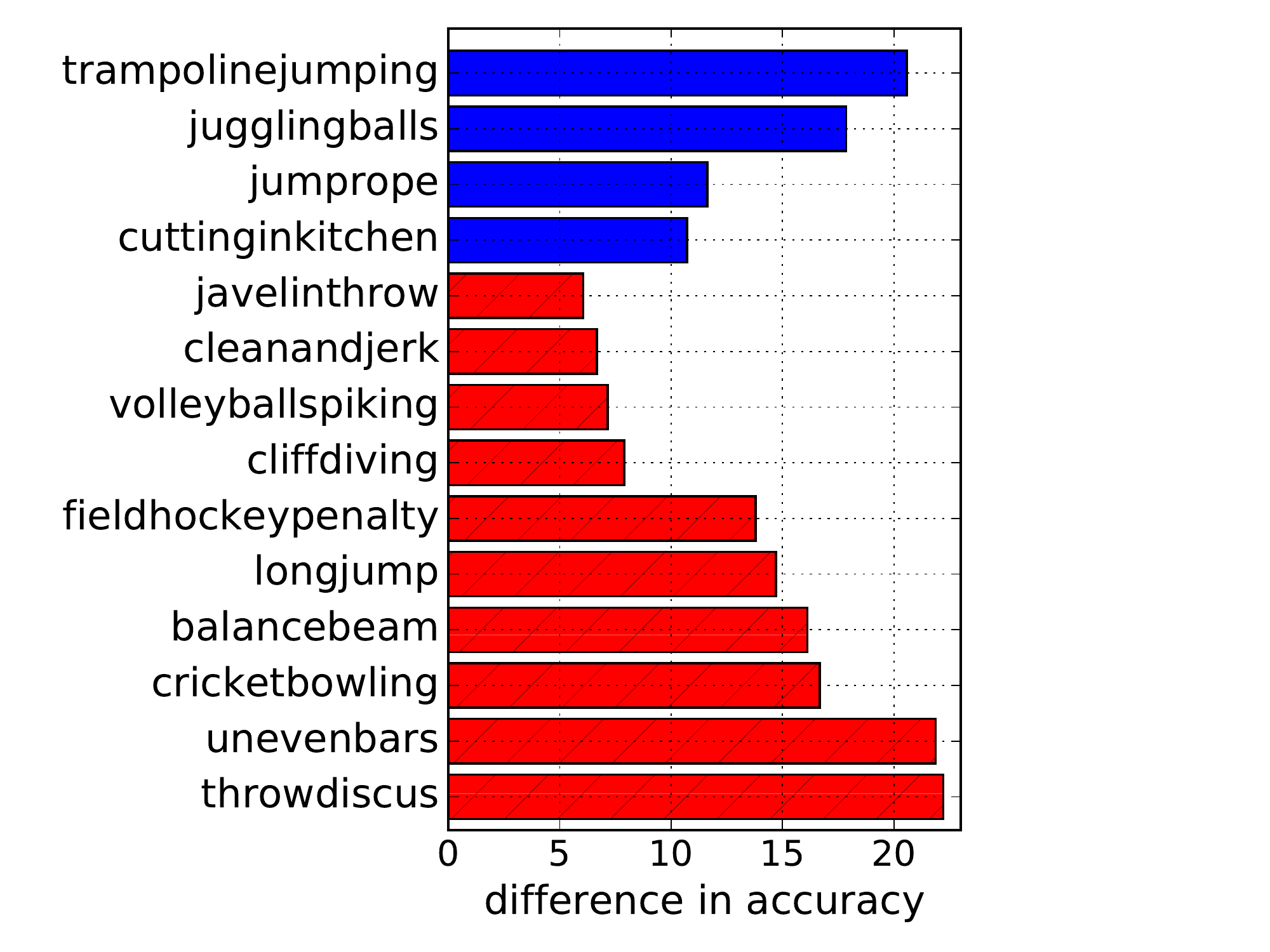}
    }
    \caption{Comparison of \adascan with mean pooling -- example classes where mean pooling is
            better (blue, top four) and vice-versa (red, all but top four). 
             }	
	\label{figClsDiff}
\end{minipage}
\vspace{-0.5em}
\end{figure*}

\autoref{tabComp} gives the detailed comparison between the best baseline of mean pooling with the
proposed \adascan, on the two datasets UCF101 and HMDB51, as well as, the two networks, spatial and
temporal. We observe that the proposed \adascan consistently performs better in all but one case out
of the $12$ cases. In the only case where it does not improve, it does not deteriorate either. The
performance improvement is more with the UCF101 dataset, \ie $77.6$ to $78.6$ for the spatial
network and $82.4$ to $83.4$ for the temporal network, on average for the three splits of the
datasets. The improvements for the HMDB51 dataset are relatively modest, \ie $40.9$ to $41.4$ and
$48.6$ to $49.2$ respectively. Such difference in improvement is to be somewhat expected. Firstly
HMDB51 has fewer samples compared to UCF101 for training \adascan. Also, while UCF101 dataset has actions related to
sports, the HMDB51 dataset has actions from movies. Hence, while UCF101 actions are expected to have
smaller sets of discriminative frames, \eg throwing a basketball \vs just standing for it,
compared to the full videos, HMDB51 classes are expected to have the discriminative information
spread more evenly over all the frames. We could thus expect more improvements in the former case,
as observed, by eliminating non-discriminative frames \cf the later where there is not much to
discard. A similar trend can be seen in the classes that perform better with \adascan \cf mean
pooling and vice-versa (\autoref{figClsDiff}). Classes such as ``throw discuss'' and ``balance
beam'', which are expected to have the discriminative information concentrated on a few frames, do
better with \adascan while others such as ``juggling balls'' and ``jump rope'', where the action is
continuously evolving or even periodic and the information is spread out in the whole of the video,
do better with mean pooling.


\begin{figure}
	\centering
	\includegraphics[width=0.9\columnwidth, trim=0 55 -10 80,clip]{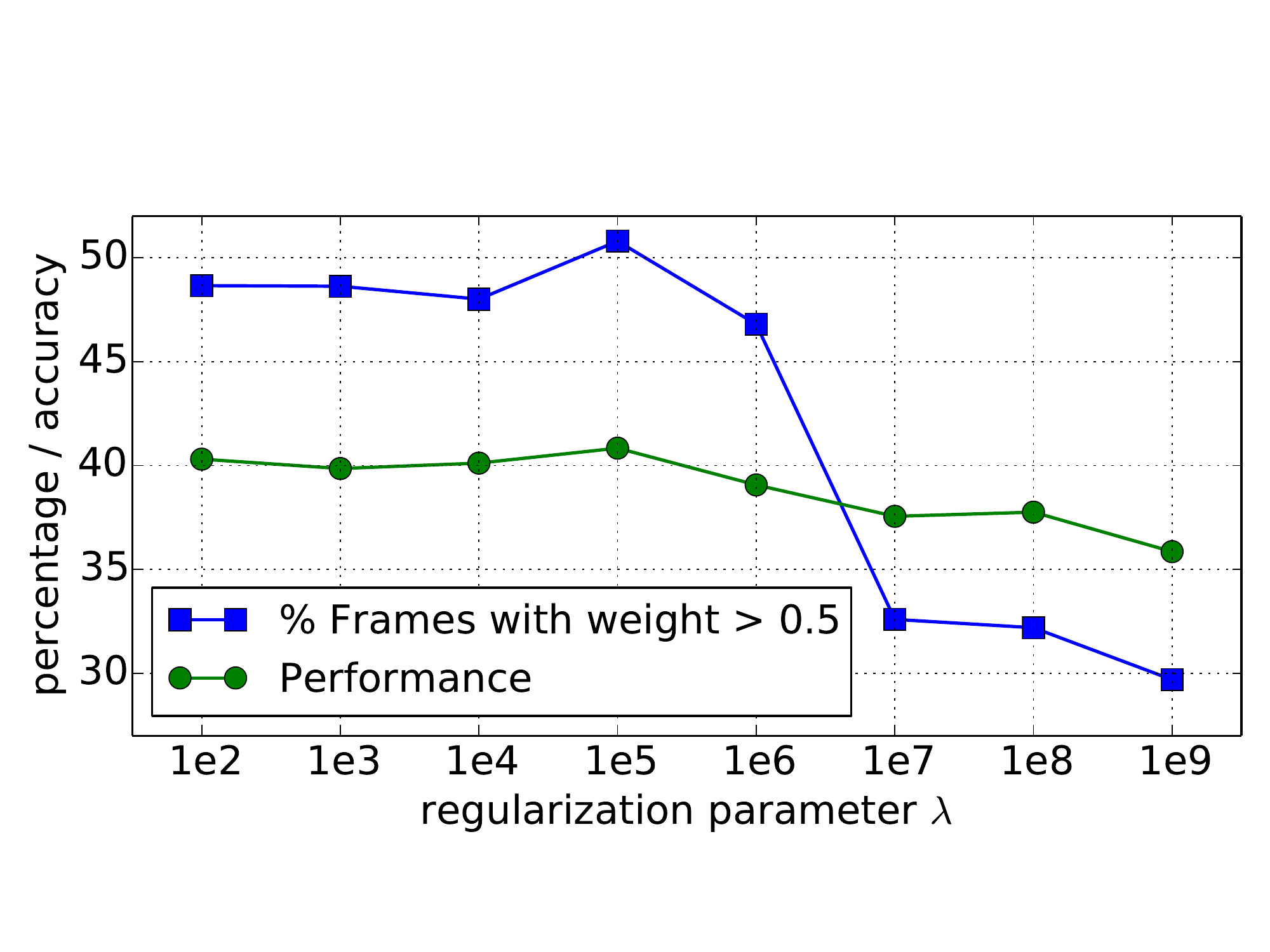}
    \caption{Effect of regularization parameter $\lambda$}	
	\label{lambda}
    \vspace{-1em}
\end{figure}

\subsubsection{Effect of Regularization Strength}
\label{secExpLambda}
As discussed in the \autoref{secLoss} above, we have a hyperparameter $\lambda \in \R^+$ which
controls the trade-off between noisy frame pruning and model fitting. We now discuss the effect of the
$\lambda$ hyperparameter. To study its effect we trained our spatial network with different $\lambda$ values 
on the HMDB51 dataset for 3 epochs to produce the shown results. We see in
\autoref{lambda} that for very low regularization ($1e2$ to $1e4$), the model gives an importance 
(\ie value of the coordinate corresponding to the frame in the normalized vector $\Gamma$ of weights) of
greater than $0.5$ to only about $50\%$ of frames, showing that the architecture in itself holds the
capability to filter out frames, probably due to the residual nature of the input to the Adaptive
Pooling module. As we increase regularization strength from $1e6$ to $1e7$ we see that we can achieve
a drastic increase in sparsity by allowing only a small drop in performance. Subsequently, there
is a constant increase in sparsity and corresponding drop in performance. The change in sparsity and
performance reduces after $1e7$ because we clip gradients over a fixed norm, thus disallowing very
high regularization gradients to flow back through the network. The $\lambda$ hyperparameter therefore
allows us to control the effective number of selected frames based on the importances predicted by the model.

\begin{figure*}
        \centering
        \vspace{-0.5em}
        \includegraphics[width=\textwidth, trim=190 360 530 215, clip]{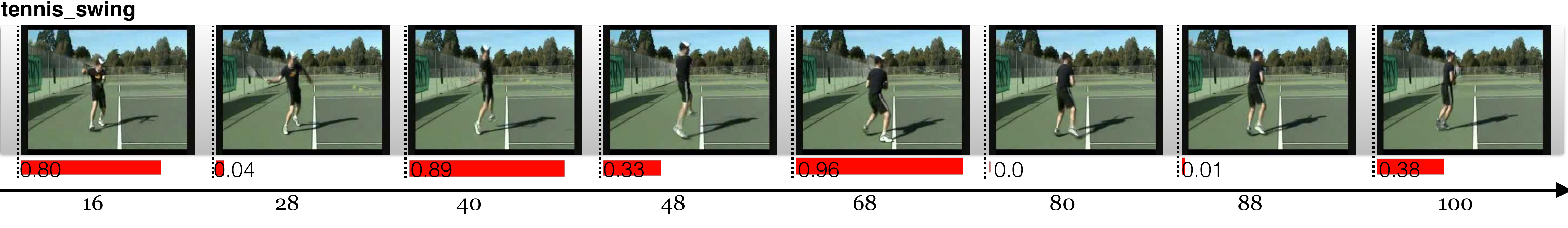}       
        \vspace{-0.5em}
        \includegraphics[width=\textwidth, trim=190 360 530 215, clip]{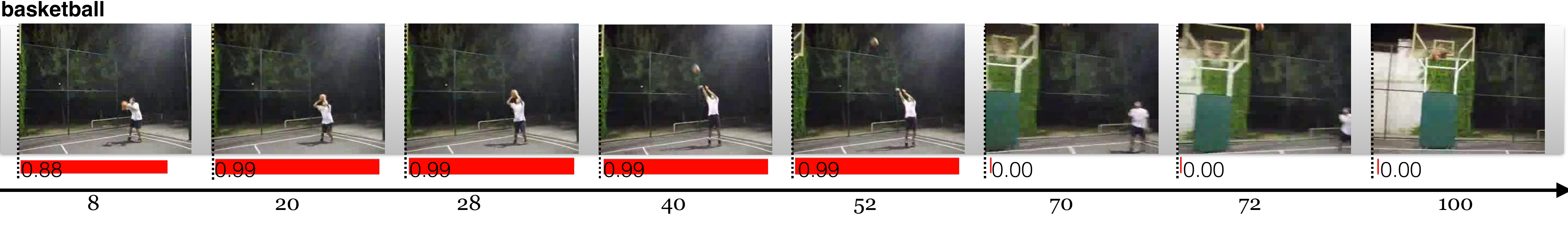}
        \vspace{-0.5em}       
        \includegraphics[width=\textwidth, trim=190 360 530 215, clip]{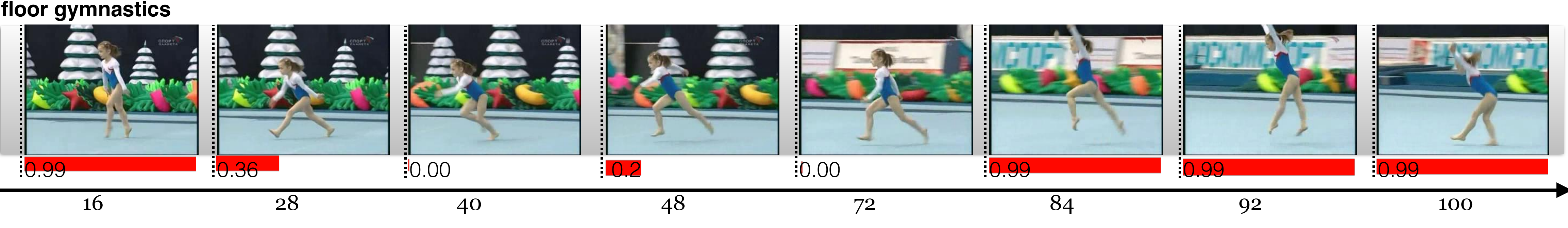}       
        \vspace{-0.5em}
        \includegraphics[width=\textwidth, trim=190 360 530 215, clip]{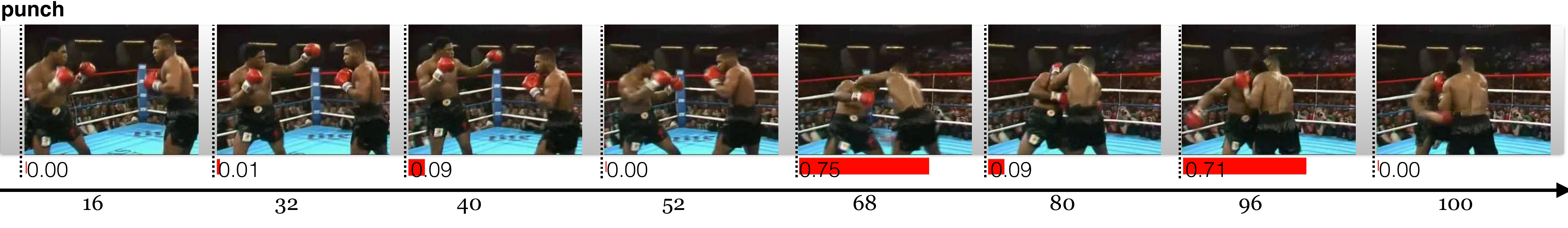}       
        \vspace{-1.5em}
        \caption{Visualizations of \adascan frame selection. The numbers and red bars below the
        frames indicate the importance weights. The timeline gives the position of the frame percentile of total number of frames in the video (best seen in
        colour).}
        \label{qual}
        \vspace{-1em}
\end{figure*}

\subsection{Comparison with State-of-the-Art}
Our model achieves performance competitive with the current state-of-the-art methods
(Table~\ref{soa1}) when combined with complementary video features on both UCF101 and HMDB51
datasets. We see that \adascan itself either outperforms or is competitive \wrt other methods
employing recurrent architectures (LSTMs) with only a single straightforward recurrent operation,
without having to employ spatial attention, \eg (on UCF101) $89.4$ for \adascan \vs $89.2, 77.0$ for
\cite{li2016videolstm, sharma2015action}, or deep recurrent architectures with significant extra
pre-training, like $88.6$ for \cite{YueCVPR2015}, demonstrating the effectiveness of the idea. We
also show improvements over traditional shallow features, \ie iDT \cite{wang2015action} and MIFS
\cite{lan2015beyond}, which is in tune with the recent trends in computer vision. Combined with
complementary iDT features the performance of \adascan increases to $91.3, 61.0$ from $89.4,54.9$,
which further goes up to $93.2, 66.9$ for the UCF101 and HMDB51 datasets respectively when combined with C3D features. These are
competitive with the existing state-of-the-art results on these datasets.

\subsection{Qualitative Results}
\label{Qualsec}

\autoref{qual} shows some typical cases (four test videos from split 1 of UCF101) visualized with
the output from the proposed \adascan algorithm. Each frame in these videos is shown with  the
discriminative importance (value of the $\gamma_t \in [0,1]$)  predicted by \adascan as a red bar on
the bottom of the frame along with the relative (percentile) location of the frame in the whole
video. In the ``basketball'' example we observe that \adascan selects the right temporal boundaries
of the action by assigning higher scores to frames containing the action. In the ``tennis--swing''
example, \adascan selects around three segments in the clip that seem to correspond to (i) movement
to reach the ball, (ii) hitting the shot and (iii) returning back to center of the court. We also
see a similar trend in the ``floor--gymnastics'' example, where \adascan selects the temporal parts
corresponding to (i) initial preparation, (ii) running and (iii) the final gymnastic act. Such frame
selections resonate with previous works that have highlighted the presence of generally $3$ atomic
actions (or actoms) in  actions classes that can be temporally decomposed into finer actions
\cite{gaidonPAMI2013}. We also see an interesting property in the ``punch'' example, where \adascan
assigns higher scores to frames where the boxers punch each other. Moreover, it assigns a moderate
score of $0.2$ to a frame where a boxer makes a failed punch attempt. We have also shown outputs on
a video (in \autoref{figIllus}) that contains ``hammer throw'' action and was downloaded from the internet. These visualizations
strengthen our claim that \adascan is able to adaptively pool frames in a video, by predicting
discriminativeness of each frame, while removing frames that are redundant or non-discriminative. We
further observe from these visualizations that \adascan also implicitly learns to decompose actions
from certain classes into simpler sub-events.

\section{Conclusion}
We presented an adaptive temporal pooling method, called \adascan, for the task of human action
recognition in videos. This was motivated by the observation that many frames are irrelevant for the recognition task as they are either redundant or non-discriminative. The proposed method addresses this, by learning to dynamically pool different frames for different
videos. It does a single temporal scan of the video and pools frames in an online fashion. The
formulation was based on predicting importance weights of the frames which determine their
contributions to the final pooled descriptor. The weight distribution was also regularized
with an entropy based regularizer which allowed us to control the sparsity of the pooling operation
which in turn helped control the overfitting of the model. We validated the method on two
challenging publicly available datasets of human actions, \ie UCF101 \cite{soomro2012ucf101} and
HMDB51 \cite{kuehneICCV11}. We showed that the method outperforms baseline pooling methods of max
pooling and mean pooling. It was also found to be better than Multiple Instance Learning (MIL) based
deep networks. We also show improvements over previous deep networks that used LSTMs with a much simpler and interpretable recurrent operation. We also showed that the intuitions for the design of the methods were largely
validated by qualitative results. Finally, in combination with complementary features, we also
showed near state-of-the-art results with the proposed method.

\section{Acknowledgements}
The authors gratefully acknowledge John Graham from Calit2, UCSD, Robert Buffington from INC,
UCSD, Vinay Namboodiri and Gaurav Pandey from IIT Kanpur for access to computational resources, Research-I foundation, IIT Kanpur  for support, and
Nvidia Corporation for donating a Titan X GPU.

{\small
\bibliographystyle{ieee}
\bibliography{biblio}
}

\end{document}